\documentclass{elsarticle}

\usepackage{moreverb,url}
\usepackage{tikz}
\usepackage{pgfplots}
\usepackage{ccg}
\usepackage[switch]{lineno}
\usepackage{array,booktabs}
\usepackage{balance}
\usepackage{listings}
\usepackage{amsmath}

\usetikzlibrary{shapes.geometric,arrows,positioning,shapes}
\usetikzlibrary{calc,positioning,arrows}

\usepackage[colorlinks,bookmarksopen,bookmarksnumbered,citecolor=red,urlcolor=red]{hyperref}

\newcommand\BibTeX{{\rmfamily B\kern-.05em \textsc{i\kern-.025em b}\kern-.08em
T\kern-.1667em\lower.7ex\hbox{E}\kern-.125emX}}

\usepackage[utf8]{inputenc}

\title{Learning and Executing Re-usable Behaviour Trees from Natural Language Instruction}

\author{Gavin Suddrey\corref{mycorrespondingauthor}}
\cortext[mycorrespondingauthor]{Corresponding author}
\ead{g.suddrey@qut.edu.au}

\author{Ben Talbot, Frederic Maire}
\address{QUT Centre for Robotics, Queensland University of Technology, Australia}
\date{May 2021}

\tikzset{
  basic/.style  = {draw, font=\sffamily, rectangle},
  composite/.style   = {basic, thin, align=center, fill=yellow!30, text width=2em},
  decorator/.style   = {basic, diamond, aspect=2, fill=yellow!30},
  leaf/.style = {basic, text width=4em, thin,align=center, fill=white},
  naked/.style = {leaf, }
  decorator/.style = {basic, thin, align=left, fill=pink!60, text width=6.5em}
}

\begin{document}

\begin{abstract}
    Domestic and service robots have the potential to transform industries such as health care and small-scale manufacturing, as well as the homes in which we live. However, due to the overwhelming variety of tasks these robots will be expected to complete, providing generic out-of-the-box solutions that meet the needs of every possible user is clearly intractable. To address this problem, robots must therefore not only be capable of learning how to complete novel tasks at run-time, but the solutions to these tasks must also be informed by the needs of the user. In this paper we demonstrate how behaviour trees, a well established control architecture in the fields of gaming and robotics, can be used in conjunction with natural language instruction to provide a robust and modular control architecture for instructing autonomous agents to learn and perform novel complex tasks. We also show how behaviour trees generated using our approach can be generalised to novel scenarios, and can be re-used in future learning episodes to create increasingly complex behaviours. We validate this work against an existing corpus of natural language instructions, demonstrate the application of our approach on both a simulated robot solving a toy problem, as well as two distinct real-world robot platforms which, respectively, complete a block sorting scenario, and a patrol scenario.
\end{abstract}
\maketitle

\section{Introduction}
Robots have the potential to be a transformative technology in industries such as disability or aged care \citep{Tapus2007}. The application of robots within these industries is limited however by the difficult task of creating robots that are not only sufficiently capable of operating within a broad range of dynamic home environments, but also meet the individualised needs of their users. Robots that are capable of learning from their environment are one such solution to resolving this problem. Historically however, the vast majority of robots that have been made commercially available have been specialist devices that are incapable of learning or being reprogrammed, and are designed to perform a single highly constrained task such as vacuuming \citep{Laird2017}; or have required expensive expert knowledge in robotics to set up and program and must be separated from people using physical barriers such as cages \citep{Pan2012, BlossRichard2016}. 

In an attempt to address this issue, robotics researchers and manufacturers have in recent years begun to look at collaborative robots, which are not only capable of working with and around people, but can also be interacted with and programmed by non-expert users \citep{Fitzgerald2013, Schmidbauer2020}. In order to make collaborative robots accessible to the general public however, mechanisms for programming these robots must be simple to use and understand. Indeed, researchers such as Azaria et al. \cite{Azaria2020} have shown that the ability for novice users to easily teach robots novel skills not only contributes to improved performance and quicker completion times, but also improves overall user engagement.

One common approach to enabling end-users to program robots has relied on graphical user interfaces such as drag-drop interfaces. These interfaces allow user builds increasingly complex behaviours by connecting graphical blocks representing either primitive behaviours such as motor or audio controls, or complex behaviours that embed groups of primitive and complex behaviours \citep{Pan2012}. However, while these approaches are generally more accessible than traditional scripting languages, they still rely on users having some knowledge in algorithms and data-flow, and can rapidly grow in complexity. Another common approach to teaching robots, that moves away from traditional programming techniques is teach-and-repeat. Using this approach a human operator uses a devices such as teaching pendants, training cuff, and other type of remote control to drive the robot along a desired trajectory, which the robot can record and later replay to perform a given task. Using this method allows a low-barrier to entry approach to performing simple tasks such as pick-and-place.

Less technical and potentially more accessible approaches to enabling end-users to program robots include Interactive Task Learning (ITL), in which a human instructor instructs the robot in how to perform novel tasks much in the same way a human would teach another human \citep{Laird2017}. ITL can exploit teaching methods including visual demonstration \citep{Mohseni2015}, as well as natural language dialogue \citep{She2014}, to enable novice users to teach robots and other types of artificial agents to complete novel tasks. Of these communication modalities natural language in particular has interested researchers as it is both a comfortable and natural means by which humans typically communicate information \citep{Tellex2020}, and its use as a mechanism for controlling artificial agents can be traced back to early pioneers such as Winograd \cite{Winograd1972}.

The utility of natural language as a means for interacting with technology such as robots has been further demonstrated in recent years by the surge in popularity of interactive personal assistants (IPAs), such as Siri, Google Assistant, and Amazon Alexa \citep{Lopatovska2019}. IPAs are capable of providing a wide variety of functions that can be initiated via spoken dialogue, including question answering, navigation, purchasing goods from online stores, and interfacing with and controlling smart technology. According to Barzilai and Rampino \cite{Barzilai2020}, the use of ``natural'' interactions with IPAs is one of the leading factors in their transitioning from novelty devices into ``disappearing technologies'', in which the device blends seamlessly in with the surrounding environment. This concept of ``disappearing technology'' provides a firm basis for justifying the need for natural interactions between human users and robots.

However, while natural language is a desirable means for teaching robots novel tasks, its use in robotics has remained limited due to a number of challenges. These challenges include ambiguities imposed by natural language such as syntactic and semantic ambiguities, as well as ambiguities arising from under-specified instructions \citep{Lauria2002}; expectations around the use of context and background knowledge \citep{Suddrey2016}; in addition to limitations imposed by the various formalisms used to represent learnt behaviours \citep{Tellex2020}. Collectively, these challenges not only make converting natural language into formal representations that a robot can then act upon extremely difficult, but also affect the choice of control architecture available to the agent. This latter point is of particular interest, as the choice of control architecture is of fundamental importance when designing autonomous agents \citep{Colledanchise2017}.

In this paper we demonstrate how behaviour trees \citep{Colledanchise2018}, a control architecture that has gained popularity in the last two decades as tool for modelling agent behaviour, provides a rich formalism for representing tasks described through natural language. Additionally, we present a system, which we call Lingua, that allows generalisation and re-use of learnt tasks that provide large degree of expressiveness, while also taking advantage of the modularity of Behaviour Trees to facilitate hierarchical learning and automatic synthesis. To summarise, the key contributions of this paper are:
\begin{enumerate}
    \item a method for mapping from natural language instructions to behaviour tree structures that can then be executed in real-time,
    \item a method for learning novel mappings between verbs and complex non-linear behaviour trees structures at run-time through situated dialogue,
    \item a method for resolving ambiguities when grounding symbols to referents in the world through iterative refinement using situated dialogue,
    \item a method for autonomously adjusting behaviour trees at run-time in order to resolve unsatisfied preconditions during task execution, 
    \item and a validation of our approach using both quantitative and qualitative assessments. 
\end{enumerate}
Importantly, all resources and source code related to this paper have been made open source\footnote{url will be provided after acceptance}.

The remainder of this paper is structured as follows: in Section~\ref{sec:nlp}, we provide a review of the state of the art in instructing robots from natural language; in Section~\ref{sec:btrees}, we introduce the behaviour tree formalism; in Section~\ref{sec:approach}, we introduce Lingua, and detail our approach to generating behaviour trees from natural language; in Section~\ref{sec:experiments}, we provide experimental results demonstrating the validity of our approach; and lastly, in Sections~\ref{sec:discussion} and~\ref{sec:conclusion} we provide a brief discussion and conclusion. 

\section{Instructing Robots and Natural Language}
\label{sec:nlp}
Processing and analysing natural language falls under the domain of Natural Language Processing, one of the oldest fields of artificial intelligence, and covers fields of study including natural language understanding, generation, and speech recognition. The earliest work in the area of natural language processing focused on the problem of machine translation, such as translating Russian to English, which only served to highlight the difficulty in handling the syntactic and semantic complexities of language \citep{Jones1994}. In the field of robotics some of the earliest work exploring the use of natural language was demonstrated by Winograd \cite{Winograd1972}, who presented an agent that was not only capable of answering questions regarding the state of the simulated block world the agent existed within, but also executing actions based on user commands. 

A fundamental problem in exploiting natural language as a learning modality is the problem of how to ground the symbols, or words, within natural language to sets of objects, actions or outcomes within the world. In their work, Steels and Kaplan \cite{Steels2000} show how an agent can be taught new symbol to object groundings using situated dialogue, in which the human would show the agent an object, and provide a symbol that describes that word. Conversely, exploring how natural language can be used to program mobile robots to explore miniature model towns, Lauria et al. \cite{Lauria2002} present an agent that given an initial set of symbol to action groundings, could hierarchically learn new groundings as compositions of previously learnt groundings. In contrast, Kollar et al. \cite{kollar2010} extract \textit{spatial description clauses}, a hierarchical structure describing components of a natural language input, which allows an agent to infer paths based on knowledge of the environment. More recently, researchers have explored how symbols, extracted from natural language can be leveraged to assist mobile robots in navigating previously unseen human-centric environments such as office spaces \citep{Talbot2020}.

Looking more specifically at the domain of learning from instruction, the problem of best formalising learnt behaviours or tasks within a control architecture remains unsolved. Approaches to representing this knowledge include graph-based approaches \citep{Dianov2016}, Petri Nets \citep{Gemignani2015}, as well as deep learning models \citep{He2016}. The importance of selecting an appropriate control structure cannot be understated. 

According to Colledanchise \cite{Colledanchise2017}, a control structure should minimally provide for the following design principles: 
\begin{enumerate}
    \item hierarchical organisation of behaviours,
    \item continual closed-loop execution,
    \item code re-usability,
    \item modular design,
    \item human readability,
    \item expressiveness,
    \item suitability for analysis, and
    \item suitability for automatic synthesis
\end{enumerate}

\noindent One of the most trivial graph-based approaches involves representing learnt tasks as sequences of discrete actions that the agent can then step through in order to achieve the desired goal. An example of this approach was demonstrated by Tenorth et al. \cite{Tenorth2010}, who propose a system in which the agent learns how to perform cooking tasks by leveraging step-by-step instructions extracted from WikiHow. Similarly, Scheutz et al. \cite{Scheutz2017} show how natural language can be used to generate action sequences directly natural language inputs provided by the user. In contrast to these approaches, which are limited to fixed sequences of actions, Meri{\c{c}}li et al. \cite{Mericcli2014} present a system that is capable of generating what they term Instructions Graphs, that are capable of leveraging branching and looping control structures. These Instruction Graphs are generated by searching for specific keywords/phrases within each user utterance, rather than evaluating the syntax of the utterance, and does not facilitate natural interactions, instead relying on the user to utter phrases such as ``end if'' to indicate block scope. 

In contrast to these approaches, Misra et al. \cite{Misra2015} argue that learnt tasks are best represented in the agent's memory as sets of goal conditions, rather than execution graphs, as they provide a more flexible mechanism for generalising learnt tasks to novel situations. To achieve this, they propose a grammar that maps directly from natural language inputs to sets of state predicates representing the desired outcome of the received utterance. Another approach that exploits goal conditions is presented by She et al. \cite{She2014}, who describe an agent that learns tasks as sets of goal conditions through situated dialogue. This is achieved by evaluating the difference in world state between the start and end of the interaction. Importantly, as these approaches avoid encoding policy information during the learning phase, each instantiation of a learnt task must be solved as a planning problem, in which the action space is limited to a set of low-level actions. While this allows the agent to generalise learnt tasks to different initial conditions, these approaches assume that all tasks can be represented as a set of goal conditions, an assumption which excludes tasks where the steps fundamental to completing the task cannot be easily captured as an outcome - such as in the case of hosting a dinner party. Additionally, as learnt tasks do not encode any policy information, the agent is prevented from reusing previous learning experiences to simplify its planning space, such that as tasks scale up in complexity the search problem can quickly become intractable.  

Approaches that fuse learning execution graphs with goal conditions, such as the work presented by Mohan and Laird \cite{Mohan2014}, demonstrate how agents can be taught novel tasks that not only allow the agent to generalise to novel situations, but also exploit policy information provided by the user to more efficiently execute the indicated task and solve planning problems during plan repair. Another example of this fused learning approach can be seen in the work presented by Suddrey et al. \cite{Suddrey2017}, which leveraged hierarchical task networks, along with traditional planning for plan repair, to learn complex tasks such as table cleaning. However, while these approaches allow for modular design and re-use they provide limited expressiveness, excluding concepts such as branching or iteration.

Differing from more classical graph-based approaches, Gemignani et al. \cite{Gemignani2015} show how Petri Nets can be used to learn parameterised task descriptions from natural language. However, their approach is limited to learning tasks as combinations of primitive actions, and does not allow for the re-use of learnt behaviours in speeding up task learning.

Translating deep learning approaches using instruction based learning to physical robots is difficult due to large number of training episodes required to converge to a solution \citep{Tellex2020}. In addition, exploiting the symbolic nature of graph representations provides a trivial mechanism for users to interrogate and revise plans \citep{Gemignani2015} when compared to deep learning methods.

In this paper, we demonstrate how behaviour trees, a graph-based control architecture, not only address the design principles outlined by Colledanchise \cite{Colledanchise2017}, but also how they can be generated from natural language instructions. 
\section{Behaviour Trees}
\label{sec:btrees}
Behaviour trees were first introduced by the gaming industry as an alternative to finite-state machines (FSM) to define the behaviour of non-player characters (NPC) in first-person shooters \citep{isla2005}. Since their inception, they have found increasing popularity within the gaming industry and are now available within every major game making tool, including Unreal Engine, CryEngine and Unity3D \citep{Martens2018}. Behaviour trees are a reactive AI formalism, and provide a number of advantages over FSMs, such as readability and scalability, while also providing an equal or greater degree of expressiveness when compared to FSMs and other control architectures such as Subsumption architectures \citep{Colledanchise2018}.

\begin{figure}[!ht]
  \centering
  \begin{tikzpicture}[sibling distance=7em]
    \node[composite] { ? }
      child { node [composite] { $\rightarrow$ } 
        child { node [leaf] {Enemy Located?} }
        child { node [leaf] {Attack Enemy} }
      }
      child { node[leaf] { Patrol } };
  \end{tikzpicture}
  \caption{A behaviour tree that causes an agent to attack if an enemy has been located, or patrol if no enemy is in sight. Refer to Section~\ref{sec:btree-types} for a detailed description of the node types used in this tree.}
  \label{fig:gametree}
\end{figure}
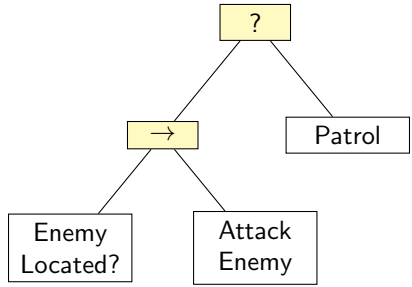

As the name implies, behaviour trees provide a mechanism for defining agent behaviour within a rooted tree structure. Formally, a behaviour tree is a directed acyclic graph $\mathcal{G}(\mathcal{V},\mathcal{E})$ containing $|\mathcal{V}|$ nodes and $|\mathcal{E}|$ arcs \citep{Martens2018}. Given a pair of nodes connected by an arc, the tail node is called the parent, and the head node the child. Each behaviour tree contains a single \textit{Root} node, which can be identified as a node with no parent, as well as one or more nodes with no children, called \textit{Leaves}. Additionally, every node within a behaviour tree falls into into one of three categories: 1) leaf nodes that represent sensing and action primitives available to the agent, 2) composite nodes that control branch selection, and 3) decorator nodes that modify branch outputs. Figure~\ref{fig:gametree} shows an example of a behaviour tree that would drive an NPC to attack enemies when they fall into the visual range of the NPC, or to patrol if no enemies are in sight. 

During execution time the agent will send a tick to the root node of the behaviour tree at a given frequency. This tick represents an activation signal and will cause the root node to then pass the tick to one of its children, depending on the type and policy of the root node. The tick is passed down the tree in this manner until it arrives at a leaf node, which immediately returns a status to its parent which will be one of three values: \textit{Success} which indicates that the leaf has completed its objective successfully; \textit{Failure}, which indicates that that the leaf has completed unsuccessfully; or \textit{Running}, which indicates that the leaf has initiated an action that will conclude during a future timestep. The return status received by the parent will then either cause the parent to tick another child, or return the received status up to its parent, depending on its type and policy. This process continues until the root node receives and returns a return status to the agent. Critically, to maintain a fixed frequency, leaf nodes in a behaviour tree should be non-blocking to allow the behaviour tree to be sufficiently reactive.

Despite their advantages, behaviour trees have only made their way into the field of robotics in recent years, with some of the earliest work by Marzinotto et al. \cite{Marzinotto2014}, who provide the first mathematical definition of behaviour trees, and address gaps in behaviour tree theory that limited its applicability to real world systems. Further work mathematically proving the robustness and safeness of behaviour trees was presented by Colledanchise and Petter \cite{Colledanchise2018}, who also demonstrated how behaviour trees can be used in a practical context to control robots that are capable of not only solving mobile pick and place challenges, but also recovering from problems encountered during task execution, such as blocked paths. 

\subsection{Node Types}
\label{sec:btree-types}
Nodes used to construct behaviour trees can be categorised into three distinct categories: 1) leaf nodes, which represent executable actions and condition checks, 2) composite nodes, that can have one or more children, and utilise specific policies on which children are to be ticked, and 3) decorator nodes which contain a single child, whose output they modify based on a node-specific policy (e.g., inverting the status outputted by the child, such that a \textit{Success} becomes a \textit{Failure} and vice versa).

\subsubsection{Leaf Nodes} are the most fundamental nodes within a behaviour tree and represent the action or sensing primitives available to the agent. Leaf nodes are typically broken into two subcategories, Action nodes which return either \textit{Success}, \textit{Failure} or \textit{Running} and are fully preemptible, allowing them to be cancelled if another branch becomes active; and Condition nodes that return either \textit{Success} or \textit{Failure}. 

One possible example of an Action leaf that would be used with a mobile robot is a Navigation Action leaf which when ticked would prompt the agent to navigate to some location in its world. While the robot is planning and/or moving, the Action leaf will return a \textit{Running} status to its parent node. If the agent arrives at the target location, the Action leaf will return a \textit{Success} status, while if the agent is unable to reach the target, because there is no viable path, or the agent has fallen over, the Action leaf will return a \textit{Failure} Status. In addition, the Action leaf can be preempted if another branch of the tree becomes active, such as in the case of a branch that defines a recharge policy that only becomes active when the battery of the robot reaches a critically low level.

\subsubsection{Composite Nodes} are nodes with one or more children where children can be of any node type. The composite node's type defines a policy which determines the child to tick, and how it handles the ticked child's return status. The two most commonly used composite nodes are \textit{Sequences} and \textit{Selectors}. 

The \textbf{Sequence} composite node uses an ordered list to hold their children. When a Sequence node is ticked, it will tick each of its children in order, returning \textit{Success} to its parent if all of its children return \textit{Success}. However, in the event a ticked child returns \textit{Running} or \textit{Failure}, the sequence node will immediately stop executing, and return the received status to its parent. In this way, Sequence nodes are logically equivalent to ordered AND operators in Boolean logic.

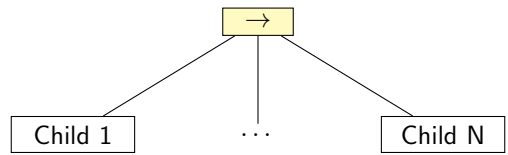
\begin{figure}[!ht]
\centering
\begin{tikzpicture}[sibling distance=7em]
  \node[composite] { $\rightarrow$ }
    child { node[leaf] { Child 1 } }
    child { node[naked] { \ldots\ } }
    child { node[leaf] { Child N } };
\end{tikzpicture}
\caption{Each child of a Sequence node must return \textit{Success} for the Sequence to return \textit{Success}.}
\end{figure}

One characteristic of \textit{Sequences} is that they are stateless. They do not hold memory regarding the return type of their children. This means that when a Sequence node is ticked, it must tick each child again, regardless of whether the child node previously returned \textit{Success} or \textit{Failure} - placing the onus on its children to remember their previous execution state. A variation of the Sequence node is the \textit{Memory Sequence}, typically denoted as Sequence*. The Sequence* node differs in that it is stateful, remembering which of its children have already been executed successfully.

The \textbf{Selector} composite node, like the Sequence node, stores its children using an ordered list which it ticks in order. What distinguishes a Selector node from a Sequence node is that a Selector node does not return \textit{Failure} upon a child returning \textit{Failure}. Instead, the Selector node will iterate through each of its children sequentially until it either finds a child that returns \textit{Success} or \textit{Running}, immediately returning the received value to its parent, or return \textit{Failure} if it exhausts its list of children. Selector nodes can therefore be considered equivalent to an ordered logical OR operator.

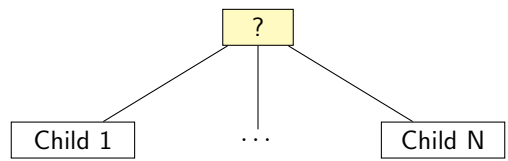
\begin{figure}[!ht]
\centering
\begin{tikzpicture}[sibling distance=7em]
  \node[composite] { ? }
    child { node[leaf] { Child 1 } }
    child { node[naked] { \ldots\ } }
    child { node[leaf] { Child N } };
\end{tikzpicture}
\caption{A Selector will return \textit{Success} if any child returns \textit{Success} or \textit{Failure} if every child returns \textit{Failure}}
\end{figure}

Variants of the Selector node may use different policies for choosing which child to tick in place of stepping through an ordered list. These include randomly selecting a child to tick, or using a priority queue. In addition to these, we also define a special case Selector, called a \textit{One-Shot Selector} that returns \textit{Success} if its set of children is empty, and in the event a child returns \textit{Success}, pops the child before returning \textit{Success} itself..

\subsubsection{Decorators} are another type of internal node. However, unlike composites, they are restricted to having a single child node. Decorators are primarily used to transform the result of their children, such as Inverter nodes, which flip \textit{Success} to \textit{Failure} and \textit{Failure} to \textit{Success}, as well as other nodes that can be used to convert from some arbitrary status to another arbitrary status. Decorators can also be used to implement behaviours such as counters and timers.

\begin{figure}[!ht]
\centering
\begin{tikzpicture}[sibling distance=7em]
  \node[decorator] { $\delta$ }
    child { node[leaf] { Child } };
\end{tikzpicture}
\caption{A Decorator, represented as a node containing a $\delta$ symbol, has only a single child whose result it transforms based on a user-defined policy.}
\end{figure}
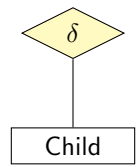

\begin{table}[!ht]
  \centering
  \caption{Additional composite and decorator node notation used within this paper}

  \begin{tabular}{>{\centering\bfseries}m{0.25\linewidth} >{\arraybackslash}m{0.65\linewidth}}
    \toprule
    Node Type & \textbf{Description} \\
    \midrule
    \tikz{\node[composite] { $\rightarrow^*$ }} & Memory Sequence node - all children must return \textit{Success} for the sequence to return \textit{Success}, does not re-tick previously succeeded children. \\ \midrule

    \tikz{\node[decorator] { $!$ }} & Inverter decorator - returns \textit{Success} if its child returns \textit{Failure} and vice versa \\ \midrule
    \tikz{\node[decorator] { $S=R$ }} & Success is Running decorator - returns either the \textit{Running} status if its child returns \textit{Success}, or the unmodified status received from its child. \\\midrule
    \tikz{\node[composite] { $\rightarrow_P$ }} & Precondition node - Please see Section \ref{sec:resolution} for a detailed description. \\ 
    \midrule
    \tikz{\node[composite] { $?^\tau$ }} & Resolution node - Please see Section \ref{sec:resolution} for a detailed description. \\ \midrule
    \tikz{\node[composite] { $\Gamma$ }} & Policy node - Please see Section \ref{sec:resolution} for a detailed description. \\ 

    \bottomrule
  \end{tabular}
\end{table}

\subsection{The Node Lifecyle}
Common implementations of behaviour trees, such as \textit{py\_trees} also define a node life-cycle, in which each node within a behaviour tree can exist within one of four distinct states. These states are \textit{Inactive}, \textit{Activating}, \textit{Active} and \textit{Deactivating}. A node transitions from an \textit{Inactive} state to an \textit{Activating} state upon being ticked, and transitions into an \textit{Active} state when it has both completed any necessary initialisation and has either ticked a child node in the case of a composite or decorator, or has initiated some action or lookup in the case of an action or condition leaf. Nodes transition from an \textit{Active} state to an \textit{Inactive} state if they are not ticked in the current timestep, and transitions into an \textit{Inactive} state once they have completed any necessary cleanup of itself and its children.






\section{From Instructions to Behaviour Trees}
\label{sec:approach}
Behaviour trees provide a flexible set of tools for representing complex agent behaviours, including modular design and reactive closed-loop execution. However, to date, the problem of mapping from natural language instructions to behaviour trees has remained unaddressed.

\begin{figure}[h!]
  \centering
  \includegraphics[width=\linewidth]{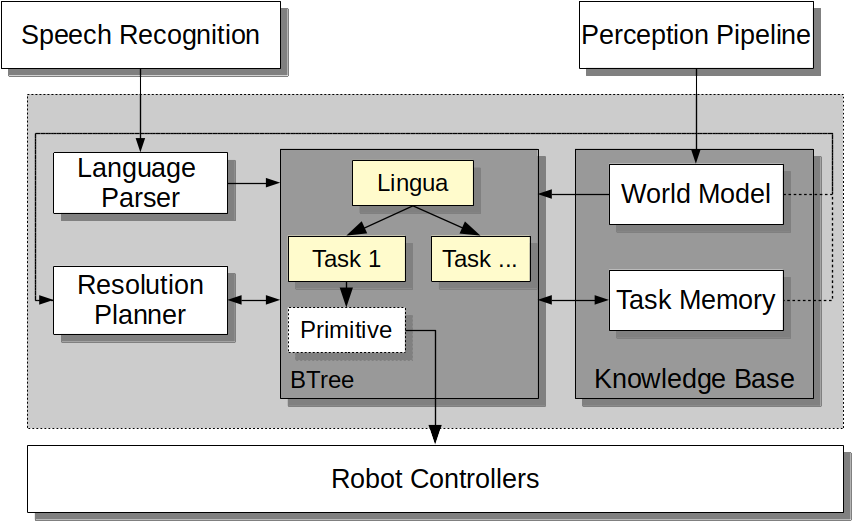}
  \caption{System Architecture of the Lingua System.}
  \label{fig:system}
\end{figure}

In this section we address this gap by describing the Lingua System (see Figure~\ref{fig:system}), which is capable of transforming natural language utterances into actionable behaviour trees that can then be executed on a robot platform. Note that the behaviour tree portions of this system (the BTree block) have been implemented using the \textit{py\_trees\_ros} and \textit{py\_trees} Python libraries.

\subsection{OpenCCG - Extracting Logical Forms from Natural Language}
Natural language input to the system is processed using the OpenCCG natural language processing library, based on the Combinatory Categorial Grammar (CCG) formalism as described by Baldridge \cite{Baldridge2002}. CCGs provide a powerful mechanism for extracting semantic information from natural language by encoding both syntactic and semantic information within the lexicon of the grammar. Words in a CCG lexicon can be grouped into two distinct classes: atomic categories, which typically include nouns and proper nouns; and functional categories, such as verbs, that consume surrounding categories as arguments. Figure~\ref{fig:entry} provides an example of an entry in a CCG lexicon, highlighting the three distinct elements of the entry: the lexical item, or word, that can appear in sentences defined by the grammar; the syntax type, which can be either atomic or function; and the semantic type, which is typically expressed using lambda calculus notation. 

\begin{figure}[!ht]
\centering
$\underbrace{\text{loves}}_{\text{Lexical Item}} \vdash\text{ } \underbrace{\underbrace{\text{S}{\backslash}\text{NP/NP}}_{\text{Syntax Type}} \text{ : } \underbrace{{\lambda} y.x.loves(x, y)}_{\text{Semantic Type}}}_\text{Category}$
\caption{The lexical entry for the verb \textit{loves} showing the key components of the entry. The syntax type indicates that the verb ``love'' generates a sentence (S) by absorbing a noun phrase (/NP) to its right, then to its left ($\backslash$NP)}
\label{fig:entry}
\end{figure}

Unlike context-free grammars, which rely on large numbers of production rules to generate a parse for some given text, the CCG formalism relies on only a small number of operators in deciding how to combine categories to create valid parses. These include: the \textit{application} operator (see Figure~\ref{fig:lambda}); the \textit{composition} operator; and the \textit{type-raise} operator. Using this small set of rules, along with the syntax types of the words in a given sentence, a CCG parser can generate logical forms that capture the semantic meaning of the sentence.  

\begin{figure}[!ht]
	\centering
  	\deriv{3}{
		{\rm Jack}&{\rm loves}&{\rm Jill}\\
		\uline{1} & \uline{1} & \uline{1} \\
		\it NP : jack  &\it S{\bs}NP/NP : {\lambda} y.x.loves(x, y) &\it NP : jill \\
		&\fapply{2}\\
		&\mc{2}{\it S{\bs}NP : {\lambda} x.loves(x, jill)} \\
		\bapply{3} \\
		\mc{3}{S : loves(jack, jill)}
	}  
  	\caption{The direction of application operator can be determined by the arrow located at the right-hand side of the operator line.}
  	\label{fig:lambda} 
\end{figure}

The set of verbs in our lexicon are extracted from the VerbNet corpus described by Kipper et al. \cite{kipper2008}. Using the associated syntax frames provided with the corpus for each verb class, the extracted verbs were assigned to one or more of the three syntactic categories, intransitive verbs (2374 verbs), transitive verbs (4399 verbs), and ditransitive verbs (3452 verbs), with a combined total of 4576 verbs across all three categories. In addition to the set of verbs in our lexicon, we also provide 8 prepositions (at, on, off, up, down, in, to, into) that can be combined with these verbs in order to construct phrasal verbs, ``look at'' or ``pick up'' that require both a verb and particle in order to obtain meaning. The set of nouns and adjectives in our lexicon is not fixed, but is contingent on both the domain in which the robot is operating, as well as the perceptual capabilities available to the robot.

The output of the OpenCCG parser is a logical form, expressed as an XML tree (see Figure~\ref{fig:xml}). This logical form encodes symbolic information on elements such as tasks that the agent should execute, and objects that the agent should act on. Additionally, control flow operators such as iterators, branches and timers, which enable the agent to learn complex non-linear behaviours, can also be described.

\begin{figure}[!ht]
\begin{lstlisting}[basicstyle=\small]
<xml>
   <lf>
      <satop nom="w0:action">
         <prop name="drop" />
         <diamond mode="tense">
            <prop name="pres" />
         </diamond>
         <diamond mode="agent">
            <nom name="x1:animate-being" />
         </diamond>
         <diamond mode="arg0">
            <nom name="w2:object" />
            <prop name="ball" />
            <diamond mode="det">
               <prop name="the" />
            </diamond>
            <diamond mode="num">
               <prop name="sg" />
            </diamond>
         </diamond>
      </satop>
   </lf>
   <target>drop the ball</target>
</xml>
\end{lstlisting}
\caption{A logical form, expressed as an XML tree, outputted by the OpenCCG parser for the instruction ``drop the ball''}
\label{fig:xml}
\end{figure}

\subsection{BTree - Learning and Executing Re-Usable Behaviour Trees}

The BTree block of our architecture (see Figure~\ref{fig:system}) is the process which is primarily responsible for building, executing, and learning behaviour trees. At its core, the BTree block is implemented as a behaviour tree built using the the py\_trees\_ros library, which contains a custom \textit{Lingua} composite node. The \textit{Lingua} composite node is a variant of the \textit{One-Shot Selector}, and plays a number of roles, including receiving parsed XML output from the OpenCCG process and transforming it into executable behaviour trees; pushing generated behaviour trees into its list of children for future execution; popping children once they have concluded successfully; as well as providing an interface that descendant nodes can use to retrieve user inputs. It should be noted the \textit{Lingua} node does not need to be placed at the root of the tree, but can be used within larger trees that incorporate safety behaviours, such as recharging when the battery is running low, or recovering from collisions; as well as fallback behaviours for when the agent would otherwise be idle.
  
\subsubsection{From Logical Forms to Behaviour Trees}\hfill

For any given utterance the OpenCCG process will typically provide the \textit{Lingua} node with a number of logical forms, expressed as XML trees, detailing potential interpretations for the utterance. For each logical form, the \textit{Lingua} node attempts to extract a \textit{semantic frame} (see Figure~\ref{fig:frame}). This semantic frame can be one of a number of types, such as a Task, Object, Conditional, Conjunction or Disjunction, as well as a Loop or Duration type. Each type will typically have an expected set of arguments, which are themselves semantic frames, allowing for the complex nesting of types such as conjunctions of tasks. During the process of extracting semantic frames, invalid semantic frames are discarded by identifying when expected arguments are missing, such as a missing arguments to a verb. If no valid semantic frames are identified, the agent will inform the user it was unable to understand the supplied utterance.

\begin{figure}[h!]
  \begin{lstlisting}[basicstyle=\small]
  task: 
    name: drop(object arg0)
    arg0:
      object:
        name: ball
        limit:
          only: 1
  \end{lstlisting}
  \caption{The semantic frame generated from the logical form in Figure~\ref{fig:xml}, for the instruction ``drop the ball''. The attribute ``limit: only: 1'' indicates that the phrase ``the ball'' should only be groundable to 1 object in the knowledge base of the robot.}
  \label{fig:frame}
\end{figure}

In the event a valid semantic frame is identified, it is then selected and is checked for the presence of anaphora (anaphora are words that derive their meaning from other expressions with a dialogue, such as the words ``he'' or ``it''). If no anaphora is located within the frame, the last task description within the semantic frame is extracted and its first argument is stored as the topic of interest. If anaphora are located, they are replaced with the previously selected topic of interest. If no topic of interest is available, the system responds by raising an error. 

The semantic frame is then converted into an executable behaviour tree structure (see Table~\ref{table:mappings}), and the relevant constructors are called for the generated tree and its children. This behaviour tree is then inserted as a child of the \textit{Lingua} node for future execution.

\subsubsection{Methods and Subtrees}\hfill

In order to generate executable behaviour trees for tasks, the system must have a mechanism for mapping from semantic frames describing tasks to the structures that describe how to complete them. To achieve this, the system generates a \textit{Subtree} node, which accepts the semantic frame as an argument, which it then inserts into the tree to be executed. This \textit{Subtree} node uses information from the semantic frame during setup to retrieve a \textit{method} from Task Memory that describes how to complete the task (see Section~\ref{sec:learning} for details on how methods are added to Task Memory). Each method in Task Memory is described as a 5-tuple $m = (L,P,E,\Pi,\Sigma)$, where $L$ is the method label used to match the method to the task (e.g. ``pick\_up(object arg0)''), $P$ is the set of preconditions for the method, E is the set of effects, $\Pi$ describes the behaviour tree that will be generated for the method, and $\Sigma$ is the argument mappings from $m$ to $\Pi$. Currently all non-primitive methods have the constraint that $P=E=\emptyset$. If a preexisting method is found for the given \textit{Subtree} the \textit{Subtree} stores the method which it will expand when it is activated. In the event that no method can be located, a \textit{LearnMethod} node is generated and inserted as the sole child of the \textit{Subtree} (see Figure~\ref{sec:learning}). 

\begin{table}[ht]
  \caption{Mappings between Semantic Types generated by the OpenCCG Parser and Behaviour Trees}
\begin{tabular}{ |l|l| } 
  \hline
  \bf Semantic Type & \bf Behaviour Tree \\ \hline
  Conditional & Selector(Inverter(condition), body) \\ \hline
  Conjunction:and & Sequence(l\_argument, r\_argument) \\ \hline
  Conjunction:or & Selector(l\_argument, r\_argument) \\ \hline
  Task & Subtree(name, arguments) \\ \hline
  Object & GroundObject(descriptors) \\ \hline
\end{tabular}
\label{table:mappings}
\end{table}

\subsubsection{Subtree Expansion}\hfill

While the Subtree node is a variant of the \textit{Sequence} composite, it differs from the \textit{Sequence} node by not maintaining a constant set of children. Instead the \textit{Subtree} node is expanded during its \textit{Activating} state to generate a set of children which are then popped during the \textit{Deactivating} state of its lifecycle. 

\begin{figure}[ht]
  \centering
  \begin{tikzpicture}[sibling distance=7em]
    \node[composite] { $?$ }
      child { node [composite] { $E$ } }
      child { node [composite] { $\rightarrow^*$ }
              child { node[composite] { $P$ } }
              child { node[leaf] { $\Pi$ } }
      };
  \end{tikzpicture}
  \caption{The children generated by a Subtree node with the attached method $m \in M$ where $E \neq \emptyset$ and $P \neq \emptyset$.}
  \label{fig:subtree}
\end{figure}
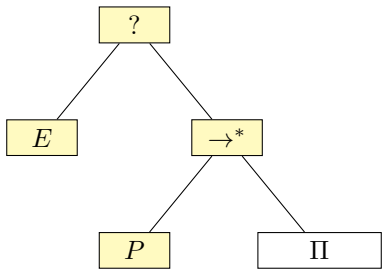

The set of children for a Subtree is generated from the elements $P$, $E$ and $\Pi$ of the method description $m$ attached to the Subtree (see Figure~\ref{fig:subtree}). Assuming that $E \neq \emptyset$ then the first child generated will be a Selector, with the set of effects on the left branch, and a subtree on the right branch. In the event the effects are already satisfied then the tree will immediately return \textit{Success} to the Subtree, otherwise the right branch of the Subtree will be ticked. If $P \neq \emptyset$, then the branch to the right of the effects will consist of a Sequence, with the set of Preconditions on the left branch, and the policy describing how to complete the task on the right.

During expansion, the set of arguments provided to the Subtree are also grounded to objects and locations within the World Model held by the agent. For each object or location, the World Model contains a mixture of continuous and discrete information captured by the perceptual pipeline of the agent. Grounding arguments to the Subtree is achieved by querying the World Model for entries based on discrete descriptors such as colour, shape type, or labels. Querying for items that satisfy these descriptors is done using a set of predicates, such as \textit{(color red ?)}, which queries for all red objects in knowledge base. In addition to visual descriptors, custom descriptors can also be implemented using custom callbacks, which can be mapped to predicates using simple pattern matching. These callbacks enable the agent to reason symbolically over continuous state information, using descriptors such as spatial relations. 

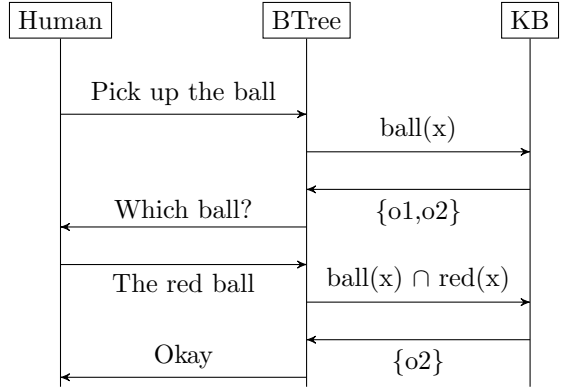
\begin{figure}[ht]
  \centering
\begin{tikzpicture}[node distance=2cm,auto,>=stealth']
  \node[draw, rectangle] (server) {BTree};
  \node[draw, rectangle, left = of server] (client) {Human};
  \node[draw, rectangle, right = of server] (kb) {KB};
  \node[below of=server, node distance=5cm] (server_ground) {};
  \node[below of=client, node distance=5cm] (client_ground) {};
  \node[below of=kb, node distance=5cm] (kb_ground) {};
  \draw (client) -- (client_ground);
  \draw (server) -- (server_ground);
  \draw (kb) -- (kb_ground);
  \draw[->] ($(client)!0.25!(client_ground)$) -- node[above,scale=1,midway]{Pick up the ball} ($(server)!0.25!(server_ground)$);
  \draw[->] ($(server)!0.35!(server_ground)$) -- node[above,scale=1,midway]{ball(x)} ($(kb)!0.35!(kb_ground)$);
  \draw[->] ($(kb)!0.45!(kb_ground)$) -- node[below,scale=1,midway]{\{o1,o2\}} ($(server)!0.45!(server_ground)$);
  \draw[<-] ($(client)!0.55!(client_ground)$) -- node[above,scale=1,midway]{Which ball?} ($(server)!0.55!(server_ground)$);
  \draw[->] ($(client)!0.65!(client_ground)$) -- node[below,scale=1,midway]{The red ball} ($(server)!0.65!(server_ground)$);
  \draw[->] ($(server)!0.75!(server_ground)$) -- node[above,scale=1,midway]{ball(x) $\cap$ red(x)} ($(kb)!0.75!(kb_ground)$);
  \draw[->] ($(kb)!0.85!(kb_ground)$) -- node[below,scale=1,midway]{\{o2\}} ($(server)!0.85!(server_ground)$);
  \draw[<-] ($(client)!0.95!(client_ground)$) -- node[above,scale=1,midway]{Okay} ($(server)!0.95!(server_ground)$);  
\end{tikzpicture}
\caption{The instruction ``pick up the ball'' indicates that we are interested in a single ball. In the event that the knowledge base (KB) contains information on two balls, one green (o1) and one red (o2), then the instruction is semantically ambiguous. Resolving this ambiguity can be done through iterative refinements provided by the user.}
\label{fig:grounding}
\end{figure}

Given an utterance from the user, we derive a set of candidate referents $X$ for each argument within the utterance, as well as an expected cardinality $n$ for $X$. If $|X| = n$ then we can ground the argument to set $X$. However, in the event that $|X| \neq n$ a Disambiguation node is inserted at the front of the Subtree node. The Disambiguation node indicates to the user that more information is needed to ground the argument, and will return \textit{Running} until the user to provide a new utterance. This new utterance is then grounded to generate a set of candidate objects $Q$, and $X$ is updated such that $X = X \cap Q$. The condition $|X| = n$ is then re-tested, with the Disambiguation node returning \textit{Success} if the condition is true, or repeating the process of requesting additional information if the condition is not satisfied. This iterative refinement process can be seen in Figure~\ref{fig:grounding}.

\subsubsection{Precondition Resolution}\hfill

\begin{figure}[ht]
\centering
\begin{tikzpicture}[sibling distance=7em]
  \node[composite] { $\rightarrow_P$ }
    child { node [leaf] { $p_1$ } }
    child { node [leaf] { $p_n$ } };
\end{tikzpicture}
\caption{The Precondition node (a subclass of the Sequence composite) with two condition nodes $p_1$ and $p_2$.}
\label{fig:P1}
\end{figure}
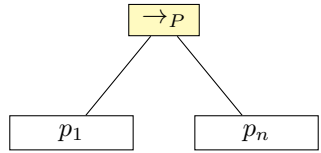

\label{sec:resolution}
The set of preconditions $P = (p_1, ..., p_n)$ for a primitive method $m$ are represented within the expanded Subtree as a set of Condition nodes underneath a Precondition node (represented in Figure~\ref{fig:P1} as a node containing $\rightarrow_P$). The Precondition node is a variant of the \textit{Sequence} and extends the functionality of the \textit{Sequence} node by allowing dynamic re-planning in the event that one or more Condition nodes are not satisfied.

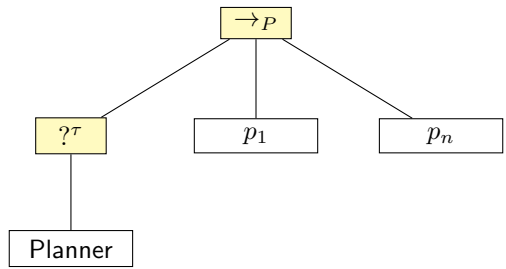
\begin{figure}[ht]
\centering
\begin{tikzpicture}[sibling distance=7em]
  \node[composite] { $\rightarrow_P$ }
    child { node [composite] { $?^\tau$ } 
        child { node[leaf] { Planner } }
    }
    child { node [leaf] { $p_1$ } }
    child { node [leaf] { $p_n$ } };
\end{tikzpicture}
\caption{When the Precondition composite node receives a \textit{Failure} status from one of the attached condition nodes (e.g., if an object to be manipulated is not reachable), a Resolution node containing a Planner action leaf is generated and prepended to the Precondition node.}
\label{fig:P2}
\end{figure}

While the Precondition node is being ticked, it will iteratively tick each Condition node, returning \textit{Success} if each and every Condition node returns \textit{Success}. However, in the event that a Condition node returns \textit{Failure}, the Precondition node responds by generating a special Resolution node, which is implemented as a \textit{One-Shot Selector} (represented in Figure~\ref{fig:P2} as a node containing $?^\tau$) containing a single Planner action leaf. This Resolution node is then prepended to the set of children of the Precondition node, after which the Precondition node returns a \textit{Running} status.

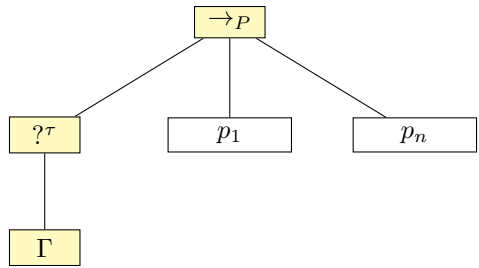
\begin{figure}[ht]
\centering
\begin{tikzpicture}[sibling distance=7em]
  \node[composite] { $\rightarrow_P$ }
    child { node [composite] { $?^\tau$ } 
        child { node[composite] { $\Gamma$ } }
    }
    child { node [leaf] { $p_1$ } }
    child { node [leaf] { $p_n$ } };
\end{tikzpicture}
\caption{The node $\Gamma$ is generated from the policy discovered by the Planner action leaf, and represents an sequence of behaviours that when completed will satisfy the conditions of the Precondition node (e.g., putting the agent in range of an object that is to be manipulated).}
\label{fig:P3}
\end{figure}

The Planner action leaf when activated will use the set of preconditions $P$ and the set of known methods $M = \{m_1, ..., m_n\}$ to find a policy $\Gamma$ such that executing $\Gamma$ will satisfy $P$. If $\Gamma$ is found, a behaviour tree is generated from $\Gamma$ which is then inserted into the Resolution node, and the Planner leaf is popped, as seen in Figure~\ref{fig:P3}. In the event that $\Gamma$ is not found, or executing $\Gamma$ does not lead to a valid resolution of $P$, then the Precondition node will respond with \textit{Failure}.

\section{Learning Behaviour Trees}
\label{sec:learning}
\begin{figure}[b!]
  \centering
  \begin{tikzpicture}[sibling distance=7em]
    \node[composite] { \textit{$\rightarrow^*$} }
      child { node [leaf] { \textit{Prompt} } }
      child { node [decorator] { $!$ }
          child { node [decorator] { $S=R$ }
              child { node [composite] { $\rightarrow^*$ }[sibling distance=6em]
                  child { node[leaf] { Poll } }
                  child { node[composite] { $?$ }[sibling distance=12em]
                      child { node[composite] { $\rightarrow^*$ }[sibling distance=6em]
                          child { node [decorator] { $!$ }
                              child { node[leaf] { Terminal? } }
                          }
                          child { node[leaf] { Push } }
                      }
                      child { node[composite] { $\rightarrow^*$ } [sibling distance=6em]
                          child { node[leaf] { Execute } }
                          child { node [decorator] { $!$ }
                              child { node[leaf] { Done? } }
                          }
                      }
                  }   
              }
          }
      }
      child { node [leaf] { \textit{Learn} } };
  \end{tikzpicture}
  \caption{The LearnMethod composite node prompts the user that it does not know how to complete the task, before entering into an interactive dialogue mode, indicated by the second branch in the tree. When the interactive dialogue is terminated by the user, the composite learns a method definition based on the dialogue.}
  \label{fig:learner}
\end{figure}
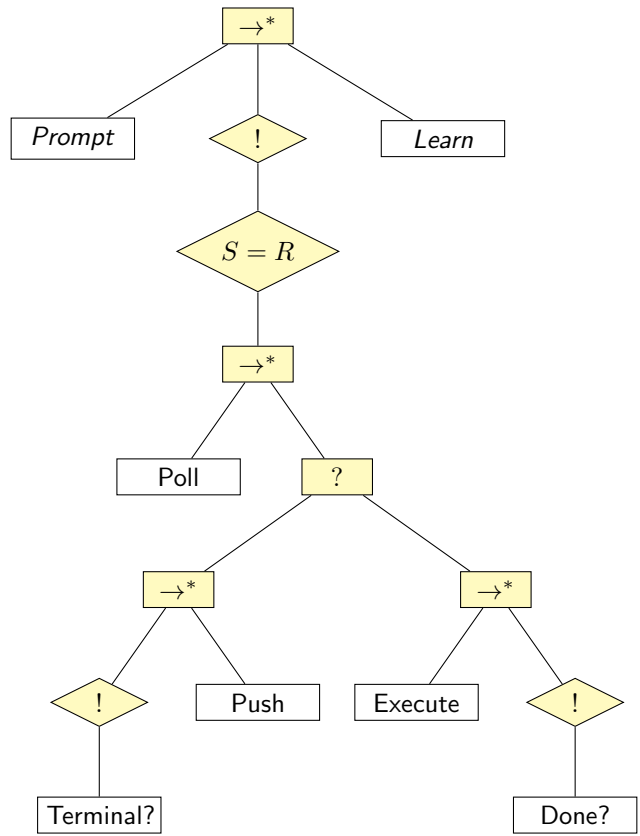

When a Subtree is setup, the Task memory of the agent is queried for a method definition that matches the task label for the Task frame assigned to the Subtree, which is then expanded when the Subtree is ticked. However, in the event that no method definition exists that corresponds to the task label, a \textit{LearnMethod} node (see Figure~\ref{fig:learner}) is generated and inserted as the sole child of the Subtree.

The \textit{LearnMethod} node is a variant of the Memory Sequence, and provides the agent with the ability to engage a human user in iterative situated natural language dialogue. When the \textit{LearnMethod} node is first ticked, it prompts the user for an explanation of the unknown task. The middle branch of the \textit{LearnMethod} node is ticked next, and continue to return a \textit{Running} status until the user indicates that the session is concluded. Within this branch, the first node to be ticked polls for a semantic frame representing either a behaviour tree to be executed, or Terminal frame, meaning that the user has indicated that the agent should stop polling for instructions. If the frame is not a Terminal, then it is inserted into a list representing the steps that the agent should execute to complete the task, and the user is polled again. If the frame is a Terminal, then the semantic frames in previously added to the list of steps are converted into behaviour trees, inserted as children into a Sequence, then ticked by the Execute node. Each item in the list is then copied into an explored list, and the user is polled to see if the task is completed. If the task is not completed, the middle branch is again ticked, and the user is prompted for the next semantic frame. If the task is completed, then the explored list is passed to the Learning node, which is then used to generate a method definition. An example dialogue between a robot and a user can be seen in Table~\ref{table:block_exchange}.

\begin{table}[ht]
    \caption{An example exchange illustrating how a user would teach a robot to put away blocks.}
  \begin{tabular*}{\linewidth}{ l  @{\extracolsep{\fill}} } 
    \hline
    H: Put the block on the table away\\
    R: How do I put the block on the table away?\\
    H: Pick it up\\
    R: Okay\\
    H: If it is red put it in the bin\\
    R: Okay\\
    H: if it is blue put it in the box\\
    R: Okay\\
    H: Try\\
    R: Let me try once\\
    R: Did I finish the task?\\
    H: Yes\\
    \hline
  \end{tabular*}
  \label{table:block_exchange}
\end{table}

\section{Experiments}
\label{sec:experiments}
To demonstrate the viability of our approach we undertook one quantitative assessment using a publicly available dataset, and three qualitative assessments using two physical robots and one simulated robot, with each robot needing to complete a robot-specific scenario. These robots are: a Franka-Emika Panda robot arm that must sort blocks when they are provided to it; an Adept Guiabot mobile platform, that must patrol between two previously unknown locations for a number of minutes; and a simulated turtlebot, that must resolve a set of unsatisfied preconditions that prevent it from completing its task. Importantly, for the sake of simplicy, text input is used in place of automatic speech recognition for all qualitative assessments undertaken in this paper. Results from our qualitative assessments can also be seen in the sumpplementary video\footnote{video url goes here}.

\subsection{Language to Behaviour Trees}

To quantitatively measure the ability of our system to generate behaviour trees from natural language, we use the VEIL-500 (Verb-Environment-Instruction-Library) version of the ``Tell-Me-Dave'' corpus \citep{Misra2016}. The VEIL-500 corpus was collected by having trial participants interact with a simulated robot in a virtual world, with participants providing sequences of instructions that the robot needed to complete. A sample of these instructions can be seen in Table~\ref{table:veil500}. In total the VEIL-500 corpus contains 2199 separate instructions, divided across 649 sets of interactions. 

\begin{table}[!ht]
  \caption{A subset of instructions found in the VEIL-500 corpus}
\begin{tabular}{ | p{0.94\linewidth} | } 
  \hline
  U: put the pot in the sink, fill it up and boil the water on the stove \\
  U: grab the vanilla syrup bottle and chocolate syrup bottle, and open the fridge door \\
  U: dump the bowl into the trash and take the bowl to the kitchen and put it in the sink \\
  U: take out food from the refrigerator, place it on a plate and then heat it in the microwave \\
  U: place the plate near the glass \\
  U: pour the kettle into the ramen \\
  U: move the cd to the television\\
  \hline
\end{tabular}
\label{table:veil500}
\end{table}

From the set of 2198 instructions we processed 2079 instructions, excluding 119 instructions that were considered invalid for reasons including non-grammatical English, spelling errors, and the use of non-imperatives. A sample of these invalid instructions can be seen in Table~\ref{table:veil500-invalid}. From these instructions a set of noun words were extracted using POS tagging, and added to the grammar used by the CCG parser.

\begin{table}[!ht]
  \caption{A subset of invalid instructions identified within  the VEIL-500 corpus}
\begin{tabular}{ | p{0.94\linewidth} | } 
  \hline
  U: put cs to game box \\
  U: -drink-food-books-paper \\
  U: tell me dave website has been visited 5000 times \\
  U: can not find drink in this scenario \\
  \hline
\end{tabular}
\label{table:veil500-invalid}
\end{table}

Each valid instruction in the corpus is initially passed through a number of preprocessing steps. These include:

\begin{enumerate}
  \item Converting the instruction to lower case
  \item Replacing simulator-specific object names with generic class names (e.g. cup1 $\rightarrow$ cup)
  \item Inserting spaces between commas and words (e.g. ``the ball, the cup'' $\rightarrow$ ``the ball , the cup'')
\end{enumerate}

The processed instructions are then fed through the OpenCCG parser to obtain a number of XML trees representing possible parses for the instruction. Each parse is then converted into a semantic frame, from which a behaviour tree is generated. 

\begin{table}[!ht]
  \caption{Examples of behaviour trees generated from instructions in the VEIL-500 corpus}
  \begin{tabular}{ |p{0.95\linewidth}| } 
  \hline
  \textbf{Instruction:} \textit{throw out the bottle, can and chips}\newline Sequence(\newline \-\hspace{0.1cm} Subtree(``throw\_out'', ``bottle''), \newline \-\hspace{0.1cm} Subtree(``throw\_out'', ``can''), \newline \-\hspace{0.1cm} Subtree(``throw\_out'', ``chips'')\newline ) \\ \hline
  \textbf{Instruction:} \textit{get food and put it on the coffee table}\newline Sequence(\newline \-\hspace{0.1cm} Subtree(``get'', ``food''),\newline \-\hspace{0.1cm} Subtree(``put\_on'', ``food'',``coffee\_table'') \newline ) \\ \hline
\end{tabular}
\label{fig:veil-btrees}
\end{table}

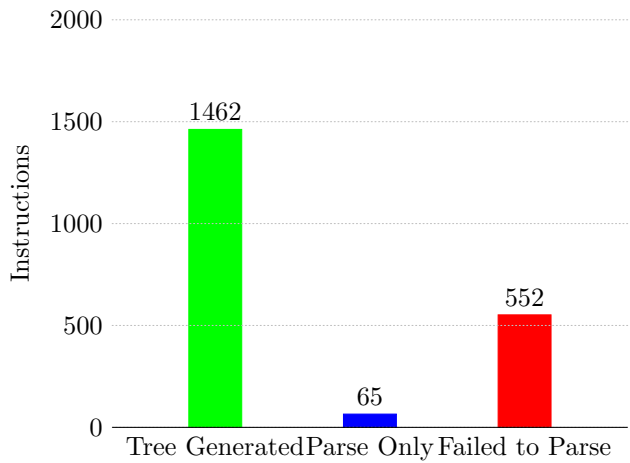
\begin{figure}[!ht]
  \centering
\begin{tikzpicture}
  \begin{axis}[
    /pgf/number format/1000 sep={},
    clip=false,
    separate axis lines,
    axis on top,
    xmin=0,
    xmax=10,
    xtick={2,5,8},
    x tick style={draw=none},
    xticklabels={Tree Generated, Parse Only, Failed to Parse},
    ymin=0,
    ymax=2000,
    ylabel={Instructions},
    every axis plot/.append style={
      ybar,
      bar width=20,
      bar shift=0pt,
      fill
    },
    axis x line*=bottom,
    axis y line*=none,
    every outer y axis line/.append style={draw=none},
    every y tick/.append style={draw=none},
    ymajorgrids,
    y grid style={densely dotted, line cap=round},
    x tick style={draw=none},
    ymin=0,
    ymax=2100,
    every node near coord/.style={
      color=black, text opacity=1
    },%
    nodes near coords
  ]
    \addplot[green] coordinates {(2, 1462)};
    \addplot[blue] coordinates {(5,  65)};
    \addplot[red] coordinates {(8,   552)};
  \end{axis}
  \end{tikzpicture}
  \caption{Summary of results from attempting to generate behaviour trees from 2079 valid natural language instructions found in the VEIL-500 corpus.}
  \label{fig:veil-results}
\end{figure}

To quantify the success of the system we categorise valid instructions based on one of three possible outcomes: 1) success, in which the instruction could be parsed by the OpenCCG grammar, and could then be converted into a behaviour tree; 2) parse, in which the instruction could be parsed by the OpenCCG grammar, but no behaviour tree could be generated, and 3) failure, in which neither a parse nor a tree could be generated. The results from this trial are shown in Figure~\ref{fig:veil-results} and a sample of the generated behaviour trees can be seen in Table~\ref{fig:veil-btrees}.

Reasons for sentences failing to be parsed included missing conjunctions between commands, such as ``get the water pour into bowl''; the use of grammatical constructs not currently covered by our grammar, such as units of measure (e.g. 1 cup of water), and sentences containing words that were missing from our grammar. The set of parses that could not be converted into behaviour trees will be missing one or more arguments while converting the parse to a semantic frame.

\subsection{Block Sorting}

In this real-robot scenario we demonstrate the ability for our system to incrementally construct generalisable behaviour trees from simple primitive behaviours, by teaching a Franka-Emika Panda robot arm to sort blocks into bins based on their respective colour.

\begin{figure}[!ht]
  \centering
  \includegraphics[width=\linewidth]{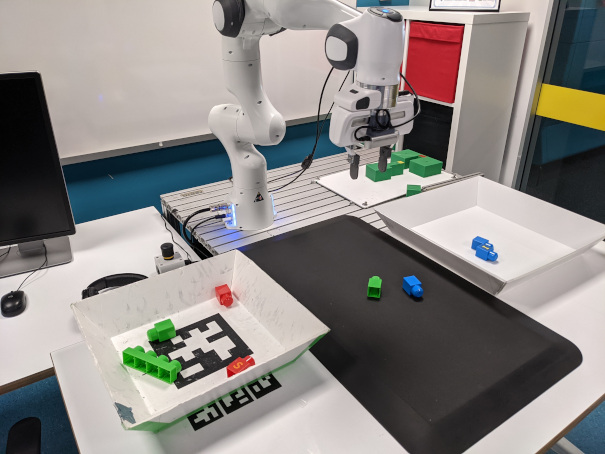}
  \caption{The workspace for the Franka-Emika Panda robot arm was comprised of three distinct locations: the table(the black mat); the bin (the left container); and the box (the right container). For simplicity, the label home was mapped to the region specified by the table. }
  \label{fig:panda}
\end{figure}

The workspace of the robot, which can be seen in Figure~\ref{fig:panda} is divided into three distinct sections, with a corresponding label and position stored in the knowledge base for each location. Initially, the robot was provided with only a simple set of primitives, a subset of which can be seen in Table~\ref{table:panda-primitives}. These primitives provide simple functionality, such as moving the end-effector to the location of objects in the workspace, and opening and closing the end-effector. Additional primitives to move the end-effector up and down were also defined but not used in this scenario. 

\begin{table}[!ht]
  \caption{The primitive methods used to complete the sort blocks scenario}
  \begin{tabular}{ |p{0.43\linewidth}|p{0.47\linewidth}| } 
  \hline
  open(tool arg0) & Opens the end-effector of the Panda robot arm \\ \hline
  close(tool arg0) & Closes the end-effector of the Panda robot arm \\ \hline
  move\_to(tool arg0, object arg1) & Moves the end-effector to the position of the specified object \\ \hline
  see(object arg0) & Tests whether an object is visible that meets the constraints defining arg0 (e.g. a red block on the table) \\ \hline
  stop() & Removes and destroys all descendant nodes of the Lingua node \\ \hline
\end{tabular}
\label{table:panda-primitives}
\end{table}

To provide sensing capabilities of its environment, the robot was equipped with a wrist-mounted D435 Realsense RGBD camera. Simple color segmentation using chromaticity was utilised to identify and classify blocks in the field of view of the robot, and visible blocks would be assigned temporary ids for the duration that they were visible to the camera.

In addition to colour, objects could also be referenced using the spatial relations ``in'' and ``on'', which were implemented using a callback that tested whether the position of objects within the knowledge base existed within an bounded region of the object used in the relation. Additional callbacks were defined for testing whether the end-effector was closed or open.

The first non-primitive taught to the Panda arm was a pick-up behaviour, with the training session being initiated with the instruction ``pick up the block''. The pick-up behaviour was described to the robot as a sequence which involved opening the robots gripper, moving to the block, closing its gripper, then moving back to its home position. Once the robot had completed a test run of this sequence, it was confirmed that it had completed the task, ending the training session and causing the agent to generate a generalised method encapsulating the process it had been guided through.

The agent was then taught a put-away behaviour, with the instruction ``put the block away''. The first step described to the robot for this behaviour was to pick up the block, re-using the behaviour learnt in the previous training session. Next, the robot was instructed to move to a location based on the colour of the block, with the robot moving to the bin if the picked block was red and green, and to the box if the picked box was blue. Lastly, the agent was instructed to open its end-effector to drop the block into the bin located below. The agent then completed a test run, the training session was ended, and a method generated.

\begin{figure}[!ht]
  \centering
  \includegraphics[width=\linewidth]{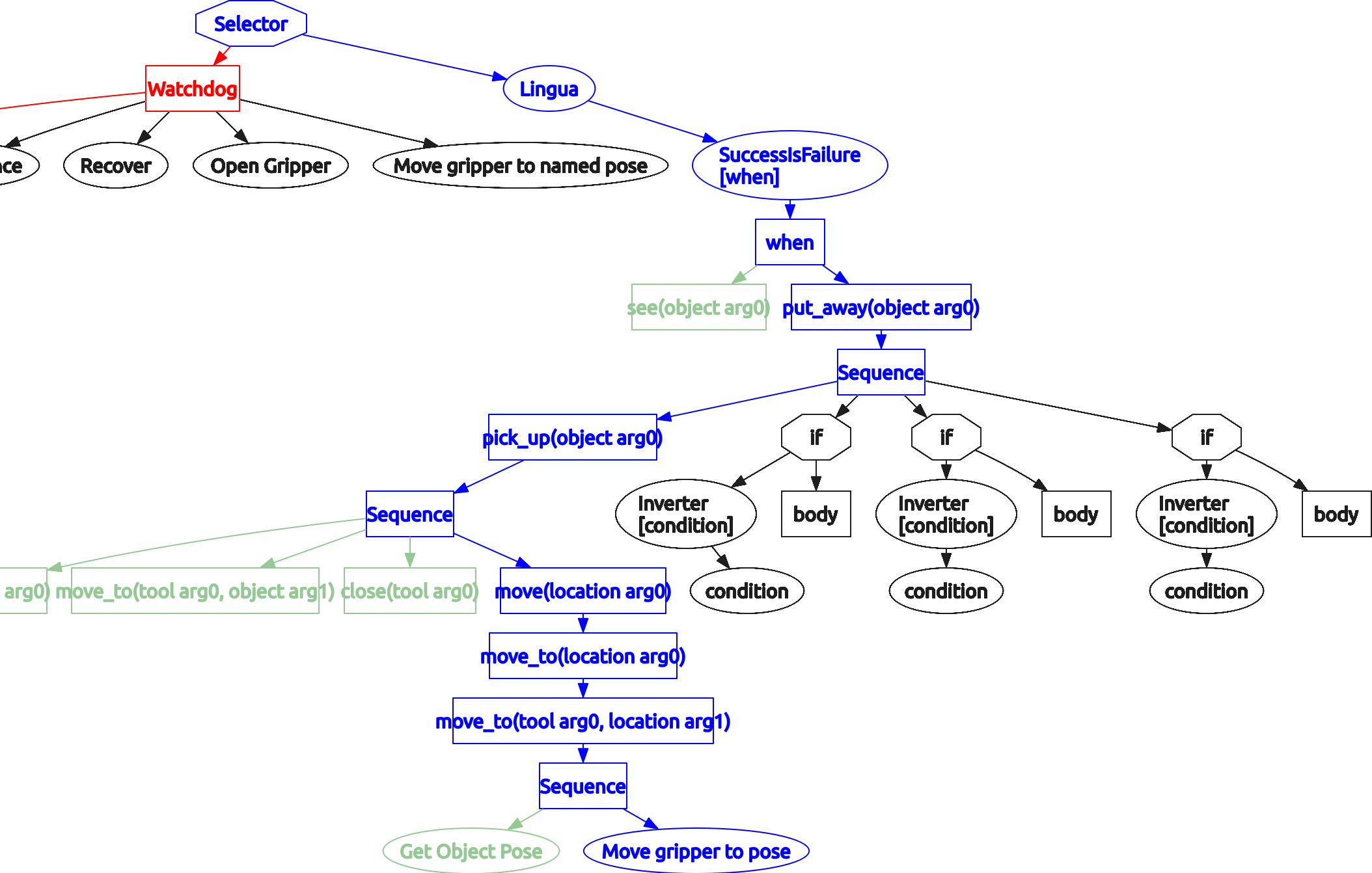}
  \caption{The behaviour tree generated from the instruction ``whenever you see a block on the table put it away'' visualised using the rqt\_py\_trees package. Rectangular non-terminals represent sequences; hexagonal non-terminals represent selectors; and oval non-terminals represent decorators. The blue path through the tree highlights the path to the currently running leaf node}
  \label{fig:panda-btree}
\end{figure}

With knowledge of how to put away blocks, the agent could then be instructed to sort blocks indefinitely with the instruction ``whenever you see a block on the table sort it''. The behaviour tree generated from this instruction can be seen in Figure~\ref{fig:panda-btree}.

\subsection{Patrolling}
\label{sec:patrol}
To further demonstrate the broad applicability of our approach, we will now describe the patrolling scenario that was completed by our Adept guiabot mobile robot. The patrolling scenario requires that we teach the mobile robot to be able to navigate between two previously unknown locations. Initially, the robot is equipped with a map of the environment, which it can localise and navigate within using the ROS navigation stack\footnote{\href{https://github.com/ros-planning/navigation}{https://github.com/ros-planning/navigation}}. The robot is also equipped with three initial primitive behaviours, described in Table~\ref{table:patrol-primitives}. In addition,  the verb ``patrol'' was added to the set of intransitive verbs previously pulled from VerbNet, as VerbNet only contained transitive and ditransitive forms.

Initially, the robot was not equipped with knowledge of any named locations within the environment. To fill in these knowledge gaps, the robot was manually driven to two separate locations, then provided with the natural language statement ``this is the kitchen/lounge''. In each case a location was stored in the knowledge base of the agent, which included a randomly generated object id, a class label (e.g. kitchen), as well as a 6-DOF pose relative to the map frame of the robot.

\begin{figure}[!ht]
  \centering
  \includegraphics[width=\linewidth]{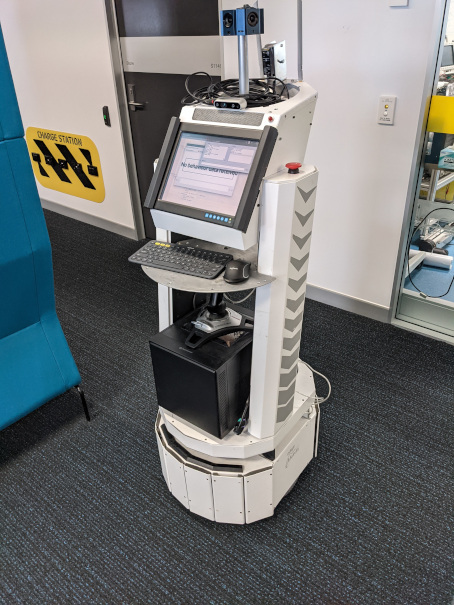}
  \caption{The Adept guiabot mobile platform was equipped with a touchscreen interface, wireless keyboard, and onboard computing.}
  \label{fig:guiabot}
\end{figure}

Using these labeled place names allowed the user to instruct the robot to either location with statements such as ``drive to the kitchen''. It is worth noting that synonymous statements such as ``go to the kitchen'' can be learnt as high-level methods that have as their child the drive\_to(room arg0) primitive, allowing multiple verbs to map to the same underlying behaviour.

\begin{figure}[!ht]
  \centering
  \includegraphics[width=\linewidth]{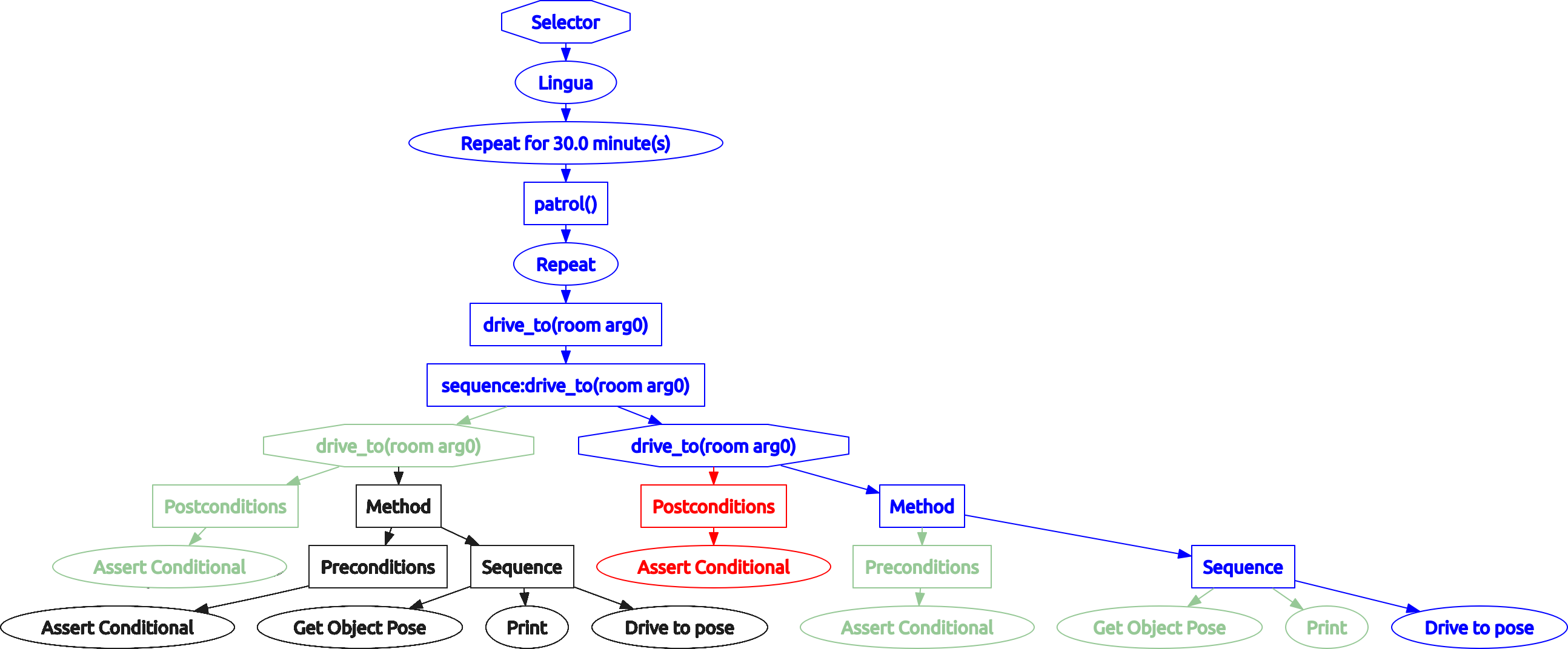}
  \caption{The behaviour tree generated from the statement ``patrol for 30 minutes'' visualised using the rqt\_py\_trees.}
  \label{fig:patrol-tree}
\end{figure}

The robot was then instructed to patrol, a task it did not know how to complete. The robot was instructed that to patrol, it should repeatedly drive between the kitchen and the lounge. The robot then performed a single lap of this trip, before asking to confirm if it had completed the patrol task successfully, and upon receiving an affirmative confirmation, ended the training session. Lastly, the agent was instructed to ``patrol for 30 minutes'', demonstrating the ability to time-bound a behaviour that would otherwise run indefinitely. The behaviour tree generated from this instruction can be seen in Figure~\ref{fig:patrol-tree}.

\begin{table}[!ht]
  \caption{The primitive methods provided for the Patrol scenario}
  \begin{tabular}{ |p{0.43\linewidth}|p{0.47\linewidth}| } 
  \hline
  is(object arg0, room arg1) & Enables the agent to store its current pose with a user-provided label (e.g. kitchen) \\ \hline
  drive\_to(room arg0) & Drives to a previously stored location based on a user-provided label (e.g. kitchen) \\ \hline
  stop() & Removes and destroys all descendant nodes of the Lingua node \\ \hline
\end{tabular}
\label{table:patrol-primitives}
\end{table}

\begin{figure}[!ht]
  \centering
  \includegraphics[width=\linewidth]{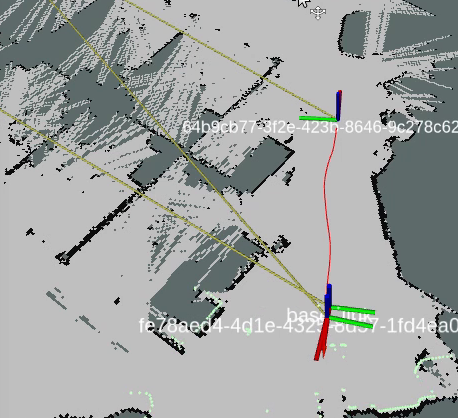}
  \caption{The map used by the mobile platform. The transform for the base link of the robot, and the two saved locations are visible.}
  \label{fig:guiabot-map}
\end{figure}

\subsection{Resolving Preconditions}

The final scenario uses a simulated case to highlight the ability of our approach to resolve unsatisfied preconditions of behaviours given incomplete instructions from a user. In this scenario, we use a turtlebot simulated using the stage robot simulator\footnote{\href{http://wiki.ros.org/turtlebot\_stage}{http://wiki.ros.org/turtlebot\_stage}}. The overall objective of this scenario is for the turtlebot to ring a bell located within its environment. The robot was equipped with the same set of primitives as the robot in Section~\ref{sec:patrol}, in addition to three extra primitives needed for the scenario. The first primitive was the ring(object arg0) primitive that completed the scenario, which had as a precondition a requirement that the robot be located near to the bell (determined by a distance threshold). The other two primitives were the \textit{lock(tool arg0)} and \textit{unlock(tool arg0)} primitives that would lock and unlock the brakes of the robot, respectively. In addition to these primitives, the driving primitives were amended to include the precondition that the brakes on the robot must had to be unlocked before the robot could move.

\begin{figure}[!ht]
  \centering
  \includegraphics[width=\linewidth]{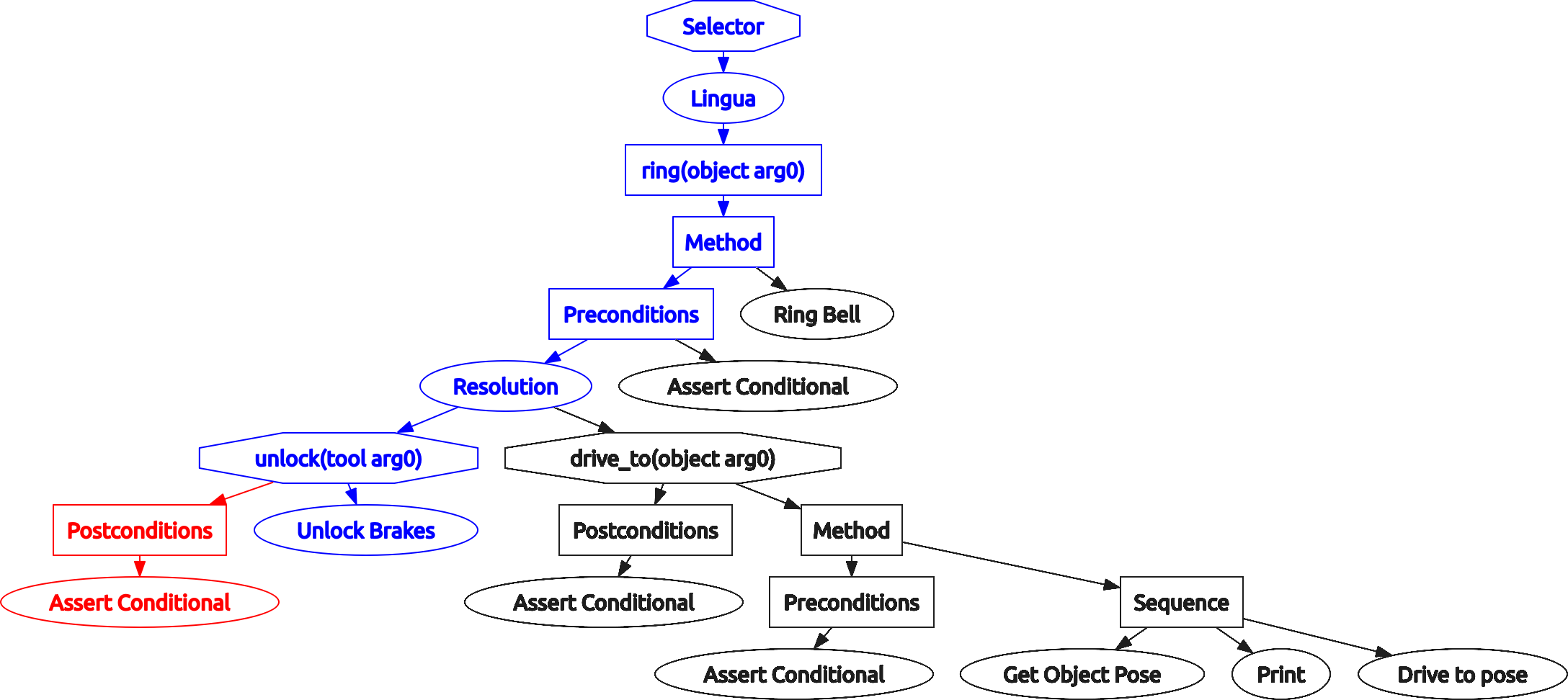}
  \caption{The Behaviour tree generated for the instruction ``ring the bell'' showing the resolution generated to resolve the precondition of the ring(object arg0) method.}
  \label{fig:stage-tree}
\end{figure}

Initially, the robot is located some distance from the bell, with its brakes locked. The robot was then prompted to ring the bell with the instruction ``ring the bell''. Importantly, the complete set of instructions needed to complete this scenario would have been ``unlock your brakes, drive to the bell, then ring the bell''. However, as the first two steps of the instruction were not provided to the agent, the agent needed to satisfy the precondition of the ring(object arg0) method selected from the provided instruction, that the robot be near the bell. Once the agent recognised that the precondition of the \textit{ring(object arg0)} primitive was unsatisfied, it responded by generating the appropriate resolver node and inserting it into the tree. This resolver node performed an iterative deepening search over the methods available to the robot, generating a plan that unlocked the brakes of the robot, before driving the robot to the bell. After completing the generated plan, the robot was then able to ring the bell and complete the scenario. A behaviour tree containing the generated solution can be seen in Figure~\ref{fig:stage-tree}. 

\section{Discussion}
\label{sec:discussion}
While in this paper we propose a method for learning generalisable behaviour trees from natural language, not all behaviours are best suited to being learnt from natural language. Rather, a complete solution to teaching and instructing everyday robots will require a combination of learning from natural language, teach and repeat, and learning from demonstration. In addition, through the use of teach-and-repeat and learning from demonstration, we can address the issue that primitive behaviours must currently be hand-coded by an expert. Robots that are able to learn how to physically execute primitives from observation, then connect these to symbols in natural language, would be powerful and flexible tools.

While we demonstrate the ability to generate behaviour trees from natural language, there are still a number of gaps in our current approach. The first of which is the absence of a mechanism for generating sets of preconditions and effects when learning novel methods. While previous approaches describe methods for generating these sets, these approaches have typically operated over linear sequences of actions \citep{Suddrey2017}. Unlike these approaches, behaviour trees introduce a number of complications in inferring preconditions and effects due largely to their non-linear nature. Future work will look at addressing this limitation.

Further, our current approach to selecting the appropriate verb type (e.g. transitive vs ditransitive) is to choose the first verb that grounds to the domain of the robot. However, this approach is not well suited to dealing with ambiguities arising from prepositional attachment. For instance, the statement ``put the ball on the table'' could be interpreted as either putting the ball onto the table (ditransitive), or putting the ball currently located on the table (transitive). While it is obvious to a native English speaker that the second sentence is nonsensical, the agent has no background knowledge from which to discard this interpretation. Approaches to deciding the most likely groundings have previously been proposed by Kollar et al. \cite{kollar2010}, and future work will look at how these approaches can be incorporated to increase the robustness of our system to grounding ambiguities.

Additionally, we do not currently make any attempts to generalise relations between the arguments supplied to behaviours being learnt, and the arguments used by their children. As such, teaching a robot to put away an object by instructing it to place the object in a specific location, will see all future objects placed in the same location. One such method to address this issue by leveraging domain knowledge available to the robot have previously been explored by Suddrey et al. \cite{Suddrey2016}, and future work will explore how this approach can be integrated into our system to resolve this limitation.

Lastly, while our approach makes use of symbolic word stems to match semantic frames to methods in our knowledge base, other representations could be used. For instance, by leveraging Word2Vec, introduced by Mikolov et al. \cite{Mikolov2013}, different verbs with similar semantic meaning, such as ``put'' or ``place'' could be matched to a single task representation by evaluating euclidean distance between their respective vectors.

\section{Conclusion}
\label{sec:conclusion}
In this paper we have outlined our approach to generated generalisable behaviour trees from natural language instruction. We have shown that natural language can be easily mapped to behaviour tree constructs, which we illustrated using the VEIL-500 dataset. Additionally, we demonstrated two real-world robot scenarios and one simulated robot scenario that illustrated the the ability of our system to generate generalisable behaviour trees that allow robots to complete such tasks as: sorting blocks based on color; patrolling between locations for fixed durations of time; and ringing bells after resolving preconditions arising from incomplete instructions provided by the user. Lastly, we have shown that the ability to generate behaviour trees from natural language instruction enables robots, and other autonomous agents, to learn novel skills from natural language instruction provided in real-time within their environment.

\section{Acknowledgements}
G.S., B.T., and F.M. acknowledge continued support from the Queensland University of Technology (QUT) through the Centre for Robotics.

\bibliographystyle{model1-num-names}

\bibliography{main}  

\end{document}